\documentclass[11pt,a4paper]{article}
\usepackage[hyperref]{emnlp2020}

\usepackage[T2A,T1]{fontenc}
\usepackage[utf8]{inputenc}
\usepackage[russian,english]{babel}

\usepackage{times}
\usepackage{latexsym}
\usepackage{tabularx}
\usepackage{hyperref}

\usepackage{microtype}

\aclfinalcopy 

\title{Reducing Unintended Identity Bias in Russian Hate Speech Detection}

\author{Nadezhda Zueva$^{1}$,\space Madina Kabirova$^{1}$,  Pavel Kalaidin$^{1, 2}$ \\
  $^1$VK, $^2$VK Lab\\
  \tt{\{firstname.lastname\}@vk.com}
  }

\date{}

\begin{document}
\maketitle
\begin{abstract}
Toxicity has become a grave problem for many online communities and has been growing across many languages, including Russian. Hate speech creates an environment of intimidation, discrimination, and may even incite some real-world violence. Both researchers and social platforms have been focused on developing models to detect toxicity in online communication for a while now. A common problem of these models is the presence of bias towards some words (e.g. woman, black, jew or \foreignlanguage{russian}{женщина, черный, еврей}) that are not toxic, but serve as triggers for the classifier due to model caveats. In this paper, we describe our efforts towards classifying hate speech in Russian, and propose simple techniques of reducing unintended bias, such as generating training data with language models using terms and words related to protected identities as context and applying word dropout to such words. 
\end{abstract}

\section{Introduction}
With the ever-growing popularity of social media, there is an immense amount of user-generated online content (e.g. as of May 2019, approximately 30,000 hours worth of videos are uploaded to YouTube every hour\footnote{\url{https://vk.cc/aANMR4}}). In particular, there has been an exponential increase in user-generated texts such as comments, blog posts, status updates, messages, forum threads, etc. The low entry threshold and relative anonymity of the Internet have resulted not only in the exchange of information and content but also in the rise of trolling, hate speech, and overall toxicity \footnote{\url{https://vk.cc/aANMZn}}.

Harassment is a pervasive issue for most online communities. A Pew survey conducted in 2014\footnote{\url{https://vk.cc/aANN6p}} found that 73\% of Internet users have witnessed online harassment, and 40\% have personally experienced it. 

Explicit policies against hate speech can be considered an industry standard\footnote{\url{https://vk.cc/aANNbQ}} across social platforms, including platforms popular among Russian-speaking users (e.g. VK, the largest social network in Russia and the CIS\footnote{\url{https://vk.cc/ayxecu}}).

The study of hate speech, in online communication in particular, has been gaining traction in Russia for a while now due to it being a prevalent issue long before the Internet \cite{lokshina2003}. The number of competitions and workshops (e.g. HASOC at FIRE-2019; TRAC 2020; HatEval and OffensEval at SemEval-2019) on the topic of hate speech and toxic language detection reflect the scale of the situation. 

Social platforms utilize a wide variety of models to detect or classify hate speech. However, the majority of existing models operate with a bias in their predictions. They tend to classify comments mentioning certain commonly harassed identities (e.g. containing words such as woman, black, jew or \foreignlanguage{russian}{женщина, черный, еврей}) as toxic, while the comment itself may lack any actual toxicity. Identity terms of frequently targeted social groups have higher toxicity scores since they are found more often in abusive and toxic comments than terms related to other social groups. If the data used to train a machine learning model is skewed towards these words, the resulting model is likely to adopt this bias\footnote{\url{https://vk.cc/aANNqT}}. 

Inappropriately high toxicity scores of terms related to specific social groups can potentially negate the benefits of using machine learning models to fight the spread of hate speech. This motivated us to work towards reducing these biases. In this paper, our main goal is to reduce the false toxicity scores of non-toxic comments that include identity terms empirically known to introduce model bias.
\section{Related Work}

\subsection{Hate Speech Detection in Russian}
Little research has been done on the automatic detection of toxicity and hate speech in the Russian language. \newcite{potapova2016} used convolutional neural networks to detect aggression in user messages on anonymous message boards. \newcite{andrusyak2018} proposed an unsupervised technique for extending the vocabulary of abusive and obscene words in Russian and Ukrainian. More recently, \newcite{smetanin2020} utilized pre-trained BERT \cite{devlin2019} and Universal Sentence Encoder \cite{yang2019} architectures to classify toxic Russian-language content.

\subsection{Reducing Unintended Bias}

\newcite{dixon2018} introduced Pinned AUC to control for unintended bias. In this paper, we adopt Generalized Mean of Bias AUCs (GMB-AUC) introduced by \cite{borkan2019}, following a study by \cite{borkan2019pinned} showing the limitations of Pinned AUC. 

\newcite{vaidya2020} proposed a model that learns to predict the toxicity of a comment, as well as the protected identities present, in order to reduce unintended bias as shown by an increase in Generalized Mean of Bias AUCs. \newcite{nozza2019} focused on misogyny detection, providing a synthetic test for evaluating bias and some mitigation strategies for it.

To our knowledge, there is no published research on reducing text classification bias in Russian.

\section{Experiments}
\subsection{Datasets}
For our experiments, we manually collected a corpus\footnote{The corpus is available on request to authors upon submitting a license agreement.} of comments posted on a major Russian social network. The mean length of each sample is 26 characters; samples over 50 characters (5\% of the total number of samples) were shortened. The corpus consists of 100,000 samples that we randomly split into training, validation and test sets in the ratio 8:1:1. Each comment was assigned a label based on whether or not it contained various forms of hate speech or abuse, including threats, harassment, insults, mentions of family members, as well as language used to promote lookism, sexism, homophobia, nationalism, etc. 

As benchmarks, we also used a small corpus of 2,000 samples in mixed Russian and Ukrainian collected by \cite{andrusyak2018}, and a corpus in Russian used by \cite{smetanin2020} (around 14,000 samples).
\subsection{Task \& Evaluation}
We considered the prediction of labels related to hate speech as a task and validated performance using introduced Generalized Mean of Bias AUCs \cite{borkan2019} to analyze whether or not the proposed methods help reduce text classification bias.

\subsection{Protected Identities}
We manually compiled a list of Russian words related to protected identities. The words were split, based on the type of hate speech used, into the following classes: lookism, sexism, nationalism, threats, harassment, homophobia, and other. Extracts from the full list are provided in Table \ref{tab:identities}. Total number of words in the list is 214. The full list of protected identities and related words is available here: \url{https://vk.cc/aAS3TQ}.

\begin{table}[!tbh]
\begin{center}
 \begin{tabularx}{\linewidth}{|X|X|} 
 \hline
 \textbf{lookism} \\
 \hline
 \foreignlanguage{russian}{корова} korova ``cow''
 \\
 \foreignlanguage{russian}{пышка} pishka ``donut (meaning "plump")''
 \\
 \hline
  \hline
 \textbf{sexism} \\
 \hline
 \foreignlanguage{russian}{женщина} zhenshchina ``woman''
 \\ 
 \foreignlanguage{russian}{баба} baba ``woman (derogatory)''
 \\
 \hline
  \hline
  \textbf{nationalism} \\
 \hline
 \foreignlanguage{russian}{чех} chekh ``"Chechen" (derogatory) lit. "Czech"''
 \\
 \foreignlanguage{russian}{еврей} evrei ``Jew''
 \\
 \hline
  \hline
   \textbf{threats} \\
 \hline
 \foreignlanguage{russian}{выезжать} vyezhat ``to come (after somebody)''
 \\
 \foreignlanguage{russian}{айпи} aipi ``ip''
 \\
 \hline
  \hline
 \textbf{harassment} \\
 \hline 
 \foreignlanguage{russian}{киска} kiska ``pussy''
 \\
 \foreignlanguage{russian}{секси} seksi ``sexy''
 \\ 
 \hline
 \hline
 \textbf{homophobia} \\
 \hline
 \foreignlanguage{russian}{гей} gay ``gay''
 \\
 \foreignlanguage{russian}{лгбт} LGBT ``LGBT''
 \\
 \hline
  \hline
 \textbf{other} \\
 \hline
 \foreignlanguage{russian}{мамка} mamka ``mother''
 \\
 \foreignlanguage{russian}{админ} admin ``admin''
 \\
 \hline
\end{tabularx}
\end{center}
\caption{Extracts from the full list of protected identities and related words.}
\label{tab:identities}
\end{table}

\subsection{Models}
We used a model based on the self-attentive encoder \cite{lin2017}. We directly feed the token embeddings matrix to the attention layer instead of the bi-LSTM encoder, making it a pure self-attention model similar to the one used in Transformer \cite{transformer}. An advantage of this architecture is that the individual attention weights for each input token can be interpretable \cite{lin2017}. This makes it possible to visualize what triggers the classifier, giving us an opportunity to explore the data and extend our list of protected identities. To overcome the problem of out-of-vocabulary words, we trained byte pair encoding \cite{bpe} on a corpora of Russian subtitles taken from a large dataset collected by \cite{taiga}, and used it for input tokenization.

We also evaluated a CNN-based text classifier (as in \cite{potapova2016}) to use as a baseline for comparison.

\subsection{Data Generation with Language Models}
To reduce model bias, we propose to extend the dataset with the output of pre-trained language models. We used the pre-trained Transformer language model\footnote{\url{https://github.com/vlarine/ruGPT2}} trained on the Taiga dataset \cite{taiga}. As Taiga contains 8 sources of normative Russian text (news, fairy tales, classic literature, etc.), we assumed that the model would be able to generate non-toxic comments even with one word from protected identities given as context. We took a random word from a list of protected identities and related words as a single word prefix for language generation, and generated samples up to 20 words long or until an end token was generated. An additional 25,000 samples were generated using the described approach and added to the existing training set.

\subsection{Identity Dropout}
Random word dropout \cite{dai2015} was shown to improve text classification. We utilized this technique to randomly (with 0.5 probability) replace protected identities in input sequences with the $<$UNK$>$ token during training.

\subsection{Multi-Task Learning}
Following \cite{vaidya2020}, we evaluated a multi-task learning framework, where we extended a base model by predicting a protected identity class from an input sequence. In our setup, the loss from an extra classifier head is weighted equal to the loss from the toxicity classifier.

\begin{table*}
\small
{\renewcommand{\arraystretch}{1.4}
\centering
\begin{tabular}{|l|c|c|c|c|c|c|}
\hline
 & \multicolumn{2}{c|}{\textbf{Our Dataset}} & \multicolumn{2}{c|}{\textbf{\cite{andrusyak2018}}} & \multicolumn{2}{c|}{\textbf{\cite{smetanin2020}}} \\ \cline{2-7} 
\textbf{Method} & GMB-AUC & F1 & GMB-AUC & F1 & GMB-AUC & F1 \\ \hline
CNN & .56$\pm$.005 & .66$\pm$.003 & .51$\pm$.005 & .59$\pm$.001&  .53$\pm$.003& .78$\pm$.002 \\ \hline
CNN + multitask & .58$\pm$.001 & .68$\pm$.008 & .52$\pm$.002 & .61$\pm$.002 & .53$\pm$.010 & .80$\pm$.002\\ \hline
Attn & .60$\pm$.002 & .71$\pm$.010 & .54$\pm$.001 & .72$\pm$.003 & .54$\pm$.005 & .80$\pm$.010 \\ \hline
Attn + multitask & .60$\pm$.004 & .74$\pm$.012 & .54$\pm$.009 & .69$\pm$.009 & .54$\pm$.007 & .82$\pm$.004 \\ \hline
Attn + LM data & .65$\pm$.003 & .74$\pm$.002 & .58$\pm$.003 & .70$\pm$.001 & .57$\pm$.006 & .83$\pm$.009 \\ \hline
Attn + LM data + multitask & .67$\pm$.002 & .74$\pm$.016 & .59$\pm$.003 & .70$\pm$.010 & .58$\pm$.003 & .84$\pm$.008 \\ \hline
Attn + identity d/o & .61$\pm$.001 & .65$\pm$.003 & .53$\pm$.004 & .68$\pm$.001 & .54$\pm$.007 & .82$\pm$.011 \\ \hline
Attn + identity d/o + multitask & .61$\pm$.005 & .66$\pm$.007 & .54$\pm$.004 & .69$\pm$.008 & .58$\pm$.009 & .83$\pm$.007 \\ \hline
Attn + identity d/o + LM data & .67$\pm$.004 & .76$\pm$.005 & .55$\pm$.003 & .71$\pm$.002 & \textbf{.59$\pm$.003} & .\textbf{86$\pm$.012} \\ \hline
Attn + identity d/o + LM data + multitask & \textbf{.68$\pm$.001} & \textbf{.78$\pm$.010} & .56$\pm$.004 & \textbf{.73$\pm$.003} & \textbf{.60$\pm$.008} & \textbf{.86$\pm$.004} \\ \hline
\end{tabular}
}
\caption{\label{citation-guide}
Generalized Mean of Bias AUCs (GMB-AUC) and F1 scores across datasets.
}
\label{tab:results}
\end{table*}

\subsection{Training Details}
We trained our models for 100,000 iterations with a batch size of 128, the Adam optimizer \cite{kingma2014adam}, and a learning rate of 1e-5 with betas (0.9, 0.999) on a single NVIDIA Tesla T4 GPU. Each experiment took approximately 1 hour to run. We used embeddings pre-trained on the corpora of Russian subtitles \cite{taiga}. We experimented with 2 different architectures (self-ATTN, CNN) in several scenarios by applying Data Generation with Language Model, Identity Dropout, and Multi-Task learning, as well as combining these approaches. We used binary cross-entropy loss as the loss function for the single-task approach. As the loss function for Multi-Task learning, we used the average loss score between two tasks: predicting the toxicity score, and predicting the protected identity class. We trained our model on the training set, controlled the training process using the validation set, and evaluated metrics on the test set. We repeated each experiment $3$ times and showed the mean and standard deviation values of the measurements. We applied an early stopping approach with patience level 50. The code is available on Google Drive\footnote{\url{https://vk.cc/aANO1g}}.

\section{Results \& Conclusion}
The results are provided in Table \ref{tab:results}.

We showed that, for our dataset and for the benchmark from \cite{smetanin2020}, adding an extra task of predicting the class of a protected identity can indeed improve the quality of toxicity classification in terms of reducing unintended bias. Moreover, we observed that simple techniques such as regularizing the input and extending the training data with external language models can help reduce unintended model bias on protected identities even further. 

For the \cite{andrusyak2018} benchmark, we did not see much improvement in our metrics. This can be attributed to language differences, as the benchmark contains abusive words both in Russian and Ukrainian.

We also observed that the proposed models achieved competitive results across all three datasets when evaluated with F1 score. The best performing model (Attn + identity d/o + LM data + multitask setup) achieved an F1 score of 0.86 on the \cite{smetanin2020} benchmark, which is 93\% of the reported SoTA performance of a much larger model fine-tuned from a BERT-like architecture.

\section{Future Work}
We are interested in automatically extending our compiled list of protected identities and related words. We also expect that fine-tuning a pre-trained BERT-like model would improve our results and plan to experiment with it.

\section{Acknowledgements}
The authors are grateful to Daniil Gavrilov and Oktai Tatanov for useful discussions, Daniil Gavrilov for review, Viktoriia Loginova and David Prince for proofreading, and anonymous reviewers for valuable comments. The authors would also like to thank the VK Moderation Team (led by Katerina Egorushkova) for their help in building a hate speech dataset.

\bibliographystyle{acl_natbib}
\bibliography{emnlp2020}







\end{document}